\documentclass[letterpaper, 10 pt, conference]{ieeeconf}  % Comment this line out if you need a4paper

\IEEEoverridecommandlockouts                              % This command is only needed if 
\usepackage{graphics} % for pdf, bitmapped graphics files
\usepackage{epsfig} % for postscript graphics files
\usepackage{amsmath}
\usepackage{amstext}
\usepackage{amssymb}
\usepackage{multirow}
\usepackage[hidelinks]{hyperref}
\usepackage{orcidlink}

\title{\LARGE \bf
DBDNet:Partial-to-Partial Point Cloud Registration with Dual Branches Decoupling
}

\author{Shiqi Li$^{\orcidlink{0000-0002-7502-896X}}$, Jihua Zhu$^{\orcidlink{0000-0002-3081-8781}}$ and Yifan Xie$^{\orcidlink{0009-0002-3225-9220}}$% <-this % stops a space
\thanks{The authors are with School of Software, Xi'an Jiaotong University, Xi'an 710000, China
and Shaanxi Joint Key Laboratory for Artifact Intelligence, China
        (e-mail {\tt\small lishiqi@stu.xjtu.edu.cn; zhujh@xjtu.edu.cn; xieyifan@stu.xjtu.edu.cn}).
}
\thanks{Jihua Zhu is the correspondence author.}
}%

\begin{document}

\maketitle
\thispagestyle{empty}
\pagestyle{empty}

%%%%%%%%%%%%%%%%%%%%%%%%%%%%%%%%%%%%%%%%%%%%%%%%%%%%%%%%%%%%%%%%%%%%%%%%%%%%%%%%
\begin{abstract}
    Point cloud registration plays a crucial role in various computer vision tasks, and usually demands the resolution of partial overlap registration in practice. Most existing methods perform a serial calculation of rotation and translation, while jointly predicting overlap during registration, this coupling tends to degenerate the registration performance. In this paper, we propose an effective registration method with dual branches decoupling for partial-to-partial registration, dubbed as DBDNet. Specifically, we introduce a dual branches structure to eliminate mutual interference error between rotation and translation by separately creating two individual correspondence matrices. For partial-to-partial registration, we consider overlap prediction as a preordering task before the registration procedure. Accordingly, we present an overlap predictor that benefits from explicit feature interaction, which is achieved by the powerful attention mechanism to accurately predict pointwise masks. Furthermore, we design a multi-resolution feature extraction network to capture both local and global patterns thus enhancing both overlap prediction and registration module. Experimental results on both synthetic and real datasets validate the effectiveness of our proposed method.
\end{abstract}


%%%%%%%%%%%%%%%%%%%%%%%%%%%%%%%%%%%%%%%%%%%%%%%%%%%%%%%%%%%%%%%%%%%%%%%%%%%%%%%%
\section{INTRODUCTION}

Point cloud registration is a fundamental problem in 3D computer vision and computer graphics. It finds wide application in various domains such as autonomous driving \cite{arnold2021fast}, robotics \cite{zhou2021path}, and augmented reality \cite{azuma1997survey}. The primary goal of this task is to estimate a rigid transformation that aligns two point clouds into the same coordinate system. Typically, the given point clouds have partial overlap and include noise.

In the task of point cloud registration, if the correspondence pairs are provided, the optimal transformation can be obtained from a closed-form solution. Conversely, if the transformation is known, the correspondence can be retrieved accordingly. The joint problem gives rise to a well-known dilemma, often referred to the chicken-and-egg problem, which makes the registration task non-trivial to solve.

Iterative Closest Point (ICP) \cite{DBLP:journals/pami/BeslM92} is the most famous algorithm for the point cloud registration problem, as an Expectation-Maximization algorithm, ICP alternates between updating correspondence to the nearest point and calculating transformation from the current correspondence. However, due to its non-convexity, ICP is sensitive to initiation and easily falls into suboptimal local minima. Various enhancement algorithms \cite{censi2008icp,yang2015go} propose using some sophisticated methods during updating the correspondence or utilizing a branch-and-bound(BnB) algorithm to navigate past the local minima. These methods are usually slower than the vanilla ICP and an acceptable result is still not guaranteed.

In recent years, deep learning-based algorithms have dominated advancements in point cloud registration. Deep Closest Point (DCP) \cite{wang2019deep} computes point correspondences via a neural network. PRNet \cite{wang2019prnet} predicts keypoint-to-keypoint correspondence to tackle partial-to-partial registration. RPMNet \cite{yew2020rpm} introduces optimal transport to get soft assignments of the point correspondence. To achieve robust registration, GeoTransformer \cite{qin2022geometric} uses transformer \cite{vaswani2017attention} to generate superpoint correspondence. GMF \cite{huang2022gmf} leverages the multimodal structure and texture information to reject the correspondence outliers. PointNetLK \cite{aoki2019pointnetlk} and FMR \cite{huang2020feature} extract two global features from input point clouds and then directly regress the transformation based on the global features.

This paper primarily focuses on two aspects of the point cloud registration problem: partial overlap and rotation-translation decoupling. In its most basic incarnation, partial-to-partial registration is more significant in practical applications. Partial overlap may cause optimization methods stuck in local minimal or confuse the learning-based methods \cite{wang2022storm}. To facilitate the model's recognition of overlapping areas, OMNet \cite{xu2021omnet} and Predator \cite{huang2021predator} predict an overlap score. However, these methods couple overlap detection with transformation estimation, potentially leading to performance degradation in both tasks. Therefore, we consider overlap detection as a preliminary step and employ a dedicated network to identify points within the overlap region. Non-overlapping points are filtered to ensure they do not hinder subsequent registration. Rotation and translation decoupling has recently gained more attention within the point cloud registration community \cite{chen2022detarnet}. In standard orthogonal Procrustes method \cite{schonemann1966generalized}, rotation $R$ is initially determined and then used as a basis to calculate translation $t$. However, as translation $t$ depends on rotation $R$, this method inevitably leads to accumulated errors in calculating translation $t$ due to errors from previous rotation calculations \cite{walker1991estimating}. To decouple the rotation and translation, we use a dual branches structure that estimates exclusive correspondence for each. Furthermore, taking inspiration from the HRNet \cite{wang2020deep}, we design a multi-resolution network to extract and fuse information from varying scales effectively. The multi-resolution network serves as a feature extractor in both the overlap detector and registration module. Extensive experiments on object-level and scene-level datasets demonstrate the effectiveness of the proposed method.

In conclusion, our contribution can be summarized as:
\begin{itemize}
    \item We propose a dual branches structure for point cloud registration, which improves the registration result by decoupling the rotation and translation.
    \item We introduce an overlap prediction module equipped with the attention mechanism to accurately filter out the non-overlapping points.
    \item We develop a multi-resolution feature extraction network to capture both local and global patterns, which provides comprehensive information for both the overlap prediction module and registration module.
\end{itemize}

\section{RELATED WORKS}
\subsection{Partial-to-partial Registration}
Partial-to-partial registration is a more realistic setting for the point cloud registration problem. PRNet \cite{wang2019prnet} detects common keypoints between partial views and establishes correspondence between them. MaskNet \cite{sarode2020masknet} learns a pointwise mask to select the overlapping points. RPMNet \cite{yew2020rpm} adds slack variables to promote the bijectivity of the matching during solving the optimal transportation. OMNet \cite{xu2021omnet}, Predator \cite{huang2021predator}, and REGTR \cite{yew2022regtr} predict the overlapping scores to tackle the partiality. Regiffusion \cite{xu2023point} applies a diffusion model to predict the complete object shape which improves the registration performance on low-overlap datasets. UDPReg \cite{mei2023unsupervised} predicts the distribution-level correspondences under the constraint of the mixing weights of GMMs to handle partial point cloud registration. In this work, we propose an overlap predictor that takes advantage of both implicit interaction from the Siamese structure and explicit interaction from the cross-attention mechanism.

\subsection{Rotation and Translation Decoupling}
The mutual interference between rotation and translation leads to performance degradation in point cloud registration. To eliminate the accumulative error in translation calculation, the dual number quaternion method \cite{walker1991estimating} represents rotation and translation using dual quaternions and calculates the two components through eigendecomposition. DeLiO \cite{thomas2019delio} decouples the rotation estimation completely by estimating and tracking surface normals. FINet \cite{xu2022finet} introduces a dual structure and a transformation sensitivity loss which improves the quality of regression by separating the solution spaces. DeTarNet \cite{chen2022detarnet} first calculates the translation from global feature offset and then mines the correct correspondences from point clouds containing only rotation. Different from previous work, our method aims to eliminate the accumulative error by separately calculating rotation and translation while minimizing additional computational complexity.

\subsection{Feature Extraction}
PointNet \cite{qi2017pointnet} pioneers geometric deep learning architectures on point cloud data. To capture local patterns, PointNet++ \cite{qi2017pointnet++} extends PointNet using a hierarchical structure. DGCNN \cite{wang2019dynamic} proposes the EdgeConv operation which incorporates local neighborhood information and can be stacked to learn global shape properties. PointConv \cite{wu2019pointconv} and KPConv \cite{thomas2019kpconv} construct convolution weights based on the continuous input coordinates. Point Transformer \cite{zhao2021point} designs a highly expressive self-attention layer for point cloud processing. To stack more layers, PointMLP \cite{DBLP:conf/iclr/MaQYR022} proposes a lightweight local geometric affine module that adaptively transforms the point feature in a local region. In this work, we utilize a multi-resolution architecture to introduce adequate information fusion across all scales, therefore the extracted feature contains both local and global information.

\section{Method}
\begin{figure*}[t]
    \centering
    \includegraphics[width=2\columnwidth]{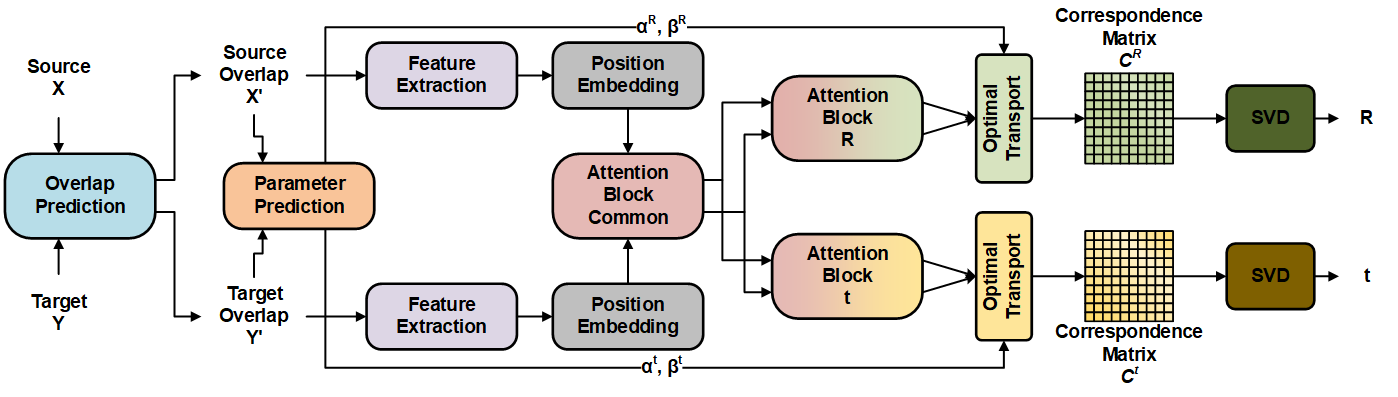} % Reduce the figure size so that it is slightly narrower than the column. Don't use precise values for figure width.This setup will avoid overfull boxes.
    \caption{Overall architecture of the proposed DBDNet. Overlapping points are first filtered by the overlap prediction module. Then, a multi-resolution network and attention blocks are used to extract and fuse pointwise features. Finally, the rotation and translation are separately calculated from dual branches.}
    \label{main}
\end{figure*}

In this section, we first introduce two basic modules: the multi-resolution feature extractor and the attention block. Then, we present the overlap prediction module to address the issue of partial overlapping. Next, we discuss the dual registration module, which aims to decouple rotation and translation. Finally, we elaborate on loss functions for training the model. Fig. \ref{main} shows the overall structure of DBDNet.

\subsection{Feature Extraction and Interaction}
\subsubsection{Multi-resolution Feature Extraction.}
\begin{figure}[t]
    \centering
    \includegraphics[width=0.95\columnwidth]{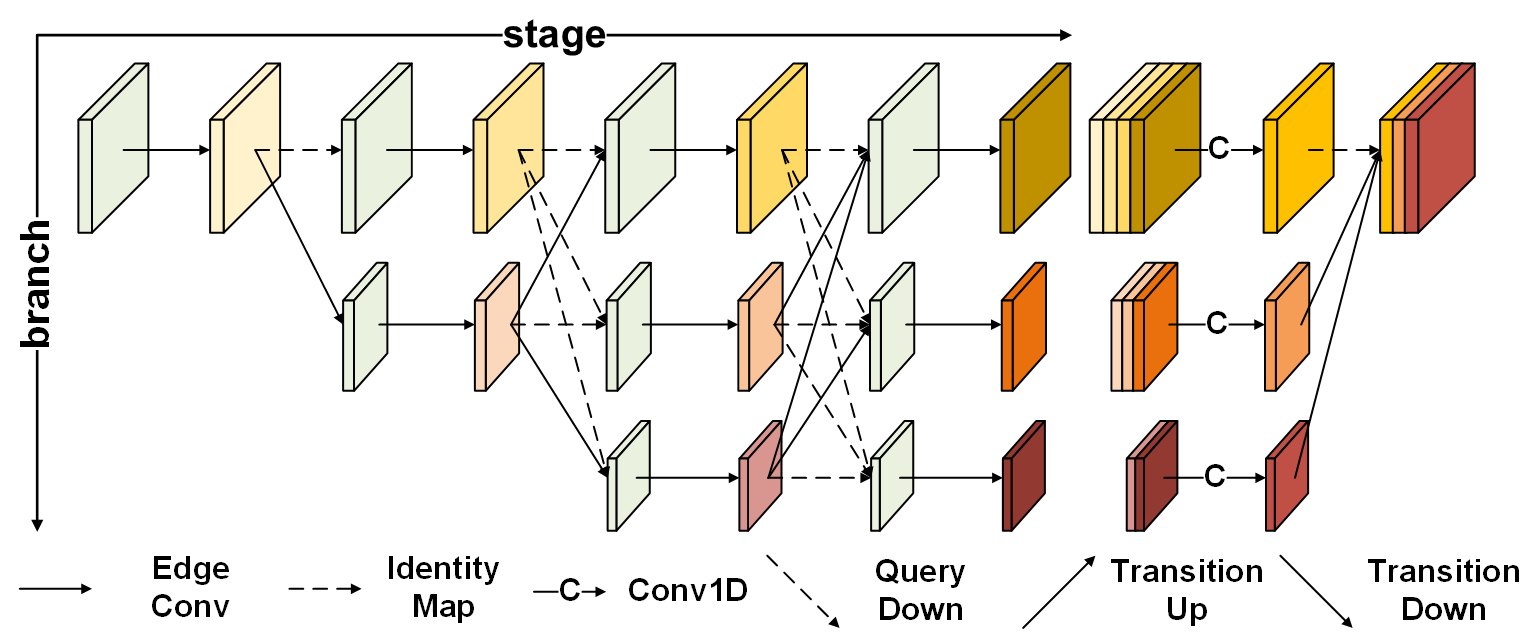} % Reduce the figure size so that it is slightly narrower than the column. Don't use precise values for figure width.This setup will avoid overfull boxes.
    \caption{Structure of multi-resolution feature extraction network.}
    \label{hr}
\end{figure}
Extracting local cues and global context is crucial for many point cloud tasks, especially in point cloud registration. Achieving accurate registration results solely based on global context is challenging, while the local feature without global information can lead to erroneous matches between locally similar structures (e.g. the legs of a chair). Inspired by HRNet \cite{wang2020deep}, we design a multi-resolution point cloud network that can extract and integrate features from different scales. The overall architecture is depicted in Fig. \ref{hr}. Our network consists of multiple parallel branches, which can be further divided into different stages. The features in different branches correspond to different downsampled point cloud subsets, while the features of the same branch capture various semantic information of the corresponding point cloud.

We employ a transition-down operation to generate the point subset and its corresponding feature for a new branch. Given the points $\mathcal{P}^l$ and features $\mathcal{F}^l$ in branch $l$. FPS (Farthest Point Sampling) is first adopted to downsample the point number by selecting half of the points from $\mathcal{P}^l$,
\begin{equation}
    \left\{\mathcal{P}^{l+1}, \widetilde{\mathcal{F}}^{l+1}\right\}= \text{FPS} \left(\left\{\mathcal{P}^{l}, \mathcal{F}^{l}\right\}\right).
\end{equation}
After that, the $k$-NN is used to group $k$ neighbors $\mathcal{N}_c$ for each point $p_c$ in $\mathcal{P}^{l+1}$ from the original points $\mathcal{P}^l$.
% \begin{equation}
%     \mathcal{N}_c = k\text{-NN}\left(p_c, \mathcal{P}^l\right).
% \end{equation}

With the indices of $k$ nearest neighbors, we can retrieve the feature in each local region and use a learnable layer $\Phi$ to extract the local pattern. Finally, the geometric information in each neighborhood region is aggregated by a symmetric function $\square$ (e.g. max),
\begin{equation}
    f_c^{l+1} = \mathop{\square}_{j\in \mathcal{N}_c } \Phi(\widetilde{f}^l_c,f_j^l).
\end{equation}

Having the points and features in each branch, we use an EdgeConv to update the features in every state. The update process in stage $s$ branch $l$ can be formulated as:
\begin{equation}
    ^{s}\widehat{\mathcal{F}}^l=\text{EdgeConv}(^s\mathcal{F}^l),
\end{equation}
where the $^s\mathcal{F}^l$ is the input feature of stage $s$ in branch $l$ and $^{s}\widehat{\mathcal{F}}^l$ is the output of stage $s$.

The repeated feature fusion across all branches enables each branch continuously receive information from other branches, therefore the feature in various scales can adequately interact. Unlike the 2D feature for images, the cross-branch feature exchange for point cloud cannot simply use strided convolution or transpose convolution layer. To plausibly transfer the feature from the upper branch $l^-$ to the lower branch $l^+$, we introduce a query-down operation. We use the indices $\mathcal{I}_{l^+}$ of points in the lower branch to retrieve the corresponding pointwise feature from the upper branch, and a learnable layer $\Upsilon $ is added to adjust the fetched feature. We formulate the query-down operation as:
\begin{equation}
\mathcal{F}^{l^+}=\{\Upsilon (f^{l^-}_i)\}_{i\in\mathcal{I}_{l^+} }.
\end{equation}

We also introduce a transition-up operation to make the information flow from the lower branch $l^+$ to the upper branch $l^-$. Since we need to propagate from sparse points to dense points, a interpolate layer is used to recover the $\mathcal{F}^{l^-}$ from $\mathcal{F}^{l^+}$, where we use inverse distance as interpolation weight \cite{qi2017pointnet++}. Similar to the query-down operation, a learnable layer $\Psi$ is also added. The operation can be formulated as:
\begin{equation}
        f_i^{l^-}=\Psi\left(\frac{\sum_{j\in\mathcal{N}_i}w_jf_j^{l^+}}{\sum_{j\in\mathcal{N}_i}w_j}\right) \quad 
        w_j=\frac{1}{\Vert p^{l^+}_i-p^{l^-}_j \Vert_2}.
\end{equation}

By utilizing the transition-up (Up) and query-down (Down) operations, we are able to facilitate information exchange throughout all branches. As a result, the process of feature fusion can be expressed as:
\begin{equation}
    ^{s+1}\mathcal{F}^l=\sum_{i=1}^{l-1}\text{Down}\left(^s\widehat{\mathcal{F}}^i\right)+^s\widehat{\mathcal{F}}^l+\sum_{i=l+1}^{L}\text{Up}\left(^s\widehat{\mathcal{F}}^i\right).
\end{equation}

To further enhance the extraction of patterns, in each branch, we concatenate output features from all stages and utilize a convolutional layer to generate the final feature \cite{huang2017densely}. Since the registration task for point clouds necessitates pointwise features, we apply the transition-up operation once again to restore the feature to its original resolution. The ultimate output of our multi-resolution network consists of the mixed feature obtained from all branches.

\subsubsection{Attention Block.}
\begin{figure}[t]
    \centering
    \includegraphics[width=0.9\columnwidth]{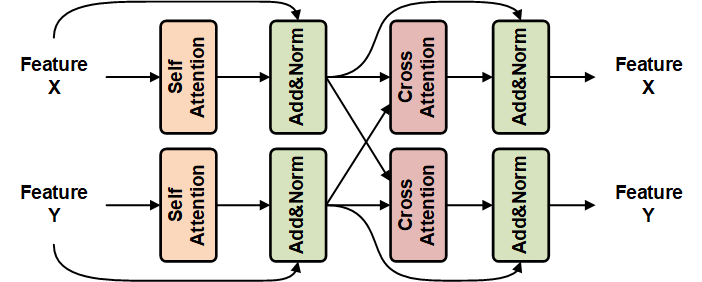} % Reduce the figure size so that it is slightly narrower than the column. Don't use precise values for figure width.This setup will avoid overfull boxes.
    \caption{Structure of attention block.}
    \label{attention}
\end{figure}
To produce abundant inter- and intra- information exchange between the source feature $\mathcal{F}_X$ and target feature $\mathcal{F}_Y$, we design an attention block. Our attention block consists of one self-attention layer and one cross-attention layer, as shown in Fig. \ref{attention}. Both two attention layers are standard Scaled Dot-Product Multi-Head Attention \cite{vaswani2017attention} which is defined as:
\begin{equation}
    \begin{split}
        \text{MHA}(Q, K, V ) = \text{Concat}\left(\text{head}_1, \cdots , \text{head}_h\right)W^O \\
        \text{head}_i = \text{Attention}\left(QW^Q_i, KW^K_i, VW^V_i\right) \\
        \text{Attention}\left(Q, K, V\right) = \text{softmax}\left(\frac{QK^T}{\sqrt{d_k}}\right)V,
    \end{split}
\end{equation}
where $W^O, W^Q_i, W^K_i, W^V_i$ are learned projection matrices, $d_k$ is the projective dimension of key.

In the self-attention layer, the query, key, and value are set to the feature from the same point cloud so that each point can acquire information from other points within the point cloud, i.e., $\mathcal{F}_{X/Y}=\text{MHA}(\mathcal{F}_{X/Y}, \mathcal{F}_{X/Y}, \mathcal{F}_{X/Y})$. In the cross-attention layer, points in one point cloud attend to points in another point cloud by changing the key and value to feature from another point cloud, i.e., $\mathcal{F}_{X/Y} =\text{MHA}(\mathcal{F}_{X/Y}, \mathcal{F}_{Y/X}, \mathcal{F}_{Y/X})$. Similar to the convincing setting, we employ a residual connection \cite{he2016deep} around each attention layer, followed by layer normalization \cite{ba2016layer}.

To emphasize the coordinates information, we employ a generalized positional encoding in 3D space \cite{yew2022regtr} before the features are fed into the aforementioned attention block. The coordination in each dimension is encoded separately, given a point located at $(x, y, z)$, the positional embedding of the $x$ dimension can be formulated as:
\begin{equation}
    \begin{split}
        \text{PE}^x[2i] &= \sin \left(\frac{x}{10000^\frac{2i}{d/3}}\right) \\
        \text{PE}^x[2i+1] &= \cos \left(\frac{x}{10000^\frac{2i}{d/3}}\right),
    \end{split}
\end{equation}
% \begin{equation}
%     \begin{split}
%         \text{PE}^x[2i] &= \sin \left(x/10000^\frac{2i}{d/3}\right) \\
%         \text{PE}^x[2i+1] &= \cos \left(x/10000^\frac{2i}{d/3}\right),
%     \end{split}
% \end{equation}
where $i$ is the index for feature dimension $d$. The positional encodings of two other dimensions are generated in the same manner, and the complete position encoding is a concatenation of three individual encodings. To make positional encodings and point features can be summed, we set the feature dimension $d$ to be divisible by 6.

\subsection{Overlap Prediction Module}

Given two point clouds, our overlap prediction module aims to predict two binary masks so that only the points in the common region of the source and target point clouds are reserved. Previous works \cite{sarode2020masknet,wu2022inenet} typically use only a Siamese network in this task, however, we argue that the Siamese structure only provides implicit information interaction between two point clouds, which is inadequate to produce a precise mask prediction. Therefore, we propose an overlap prediction that introduces explicit information exchange between source and target point clouds. 

\begin{figure}[t]
    \centering
    \includegraphics[width=0.95\columnwidth]{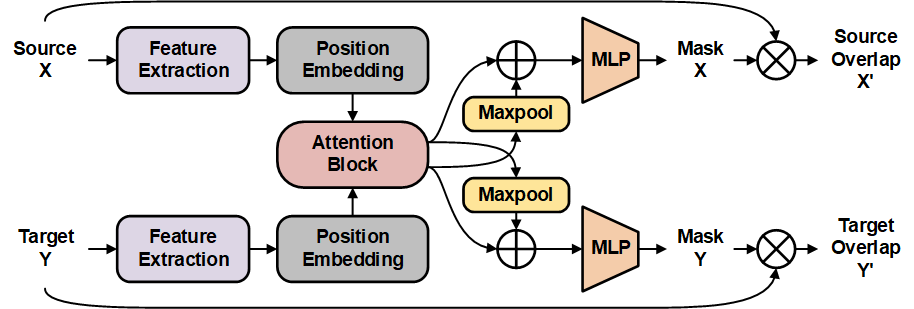} % Reduce the figure size so that it is slightly narrower than the column. Don't use precise values for figure width.This setup will avoid overfull boxes.
    \caption{Structure of overlap prediction module.}
    \label{overlapstruct}
\end{figure}

As illustrated in Fig. \ref{overlapstruct}, we first use a multi-resolution network to extract the pointwise feature from source and target points, the weights of this feature extractor are shared among two point clouds. Then stacked attention blocks are adapted to enable explicit feature interaction. After getting the interacted feature, we use max pooling to aggregate global features. The expanded global feature of the source point is concatenated to the pointwise target point cloud feature and vice verse. We use a weight-sharing MLP to predict the pointwise overlap probability $O_X\in\mathbb{R}^{|X|\times1}, O_Y\in\mathbb{R}^{|Y|\times1}$ and apply a threshold to convert the continuous probability into a binary mask. Finally, the overlap points $X'$ and $Y'$ are selected by the converted hard mask.

\subsection{Dual Branches Registration Module}

Instead of just reusing features in the overlap prediction module, we use another multi-resolution network to extract the features from filtered points. This choice is based on two considerations, on the one hand, the overlap prediction task and registration task have some commonalities but they are not identical so the feature space might be different, on the other hand, the feature from the overlap prediction module is contaminated by the points in nonoverlap region during the KNN operation, which cumber the corresponding search in the registration task.

After re-extracting the feature from overlap points, we employ stacked attention blocks to conduct feature interaction between source and target points. First, features are sent into the common attention block. Then the output features are fed into two different branches for rotation and translation decoupling. Both branches contain serval stacked attention blocks, they consist of identity structures but have different weights, thus each branch can focus on only one part of the transformation.

Having the features after rotation and translation attention blocks, we calculate the squared distance in the learned feature space to establish two initial correspondence matrixes $\widetilde{C}^R\in\mathbb{R}^{|X'|\times|Y'|}$ and $\widetilde{C}^t\in\mathbb{R}^{|X'|\times|Y'|}$:
\begin{equation}
    \widetilde{c}^{R}_{ij}=\Vert f_{x'_i}^R-f_{y'_j}^R\Vert_2^2 \quad
    \widetilde{c}^{t}_{ij}=\Vert f_{x'_i}^t-f_{y'_j}^t\Vert_2^2.
\end{equation}

Then we refine the primitive matrixes using parameters regressed by a parameter prediction module. Similar to \cite{yew2020rpm}, we utilize a PointNet to generate the parameters. The input to the PointNet consists of concatenated raw points of source and target after overlap prediction. Additionally, a tag indicating the point affiliation is included before feeding the data into the network. As we handle rotation and translation separately in two different branches, our parameter prediction module generates two distinct sets of parameters, namely $\{\alpha^R, \beta^R\}$ and $\{\alpha^t, \beta^t\}$. The adjustment can be expressed as:
\begin{equation}    
    c^{R}_{ij}=e^{-\beta^R(\widetilde{c}^{R}_{ij}-\alpha^R)} \quad
    c^{t}_{ij}=e^{-\beta^t(\widetilde{c}^{t}_{ij}-\alpha^t)}.
\end{equation}

After adjust the matrixes, we iteratively normalize across their rows and columns until the procedure converges. This operation produces two doubly stochastic matrixes \cite{sinkhorn1967concerning}. The ambiguous matches in the correspondence matrix can be suppressed during the alternate row-column two-way constraint and it greatly accelerates network optimization. Although we use an overlap prediction module to preprocess the points, there are still a small number of points that may not find a correct corresponding point in another point cloud. Therefore we also introduce slack variables, i.e., an additional row and column, to the doubly-normalization input.

Pointwise correspondence can be built from normalized matrixes easily, two corresponding relationships in rotation and translation branches are defined as:
\begin{equation}
    x'_i\Leftrightarrow \widehat{y}=\sum_{j}^{|Y'|}c_{ij}^{R/t}\cdot y'_j.
\end{equation}

We use SVD to calculate two rigid transformations from dual branches, in the rotation branch, we just calculate the $R$ matrix, and in the translation branch, we calculate its own rotation matrix and then calculate the $t$ vector. The rotation matrix in the translation branch is just an intermediate variable, and there is no loss to restrict its value. Finally, the $R$ from the rotation branch and $t$ vector from the translation branch are collected as the output.
\subsection{Loss Function}
\subsubsection{Overlap Loss}
To train the overlap prediction module, we use a binary focal loss \cite{lin2017focal} as supervision. The loss for the source point $X$ can be formulated as:
\begin{equation}
    \begin{aligned}
    \mathcal{L}_o^X=\frac{1}{|X|}\sum_{i}^{|X|} \big{(} &\alpha(1-o_i)^\gamma o_{gt_i}\log o_i+ \\
    &(1-\alpha)o_i^\gamma(1-o_{gt_i})\log(1-o_i)\big{)},
    \end{aligned}
\end{equation}
where $\alpha$ is a hyperparameter to balance overlapping and non-overlapping points, $\gamma$ is another hyperparameter to focus training on hard negatives. $o_{gt}$ is the ground true overlap label which is defined as:
\begin{equation}
    o_{gt_i}=
    \begin{cases}
        1, & \Vert \mathcal{T}_{gt}(x_i)-\text{NN}\left(\mathcal{T}_{gt}(x_i), Y\right)\Vert_2<\epsilon_o \\
        0, & \text{otherwise}
    \end{cases},
\end{equation}
where $\epsilon_o$ is overlap threshold, $\mathcal{T}_{gt}(\cdot)$ denotes the application of the ground truth rigid transform and $\text{NN}(\cdot)$ denotes the spatial nearest neighbor. The loss for the target point cloud can be calculated in the same way, the final loss is the mean of source loss and target loss, i.e., $\mathcal{L}_o=\frac{1}{2}(\mathcal{L}^X_o+\mathcal{L}^Y_o)$. 

\subsubsection{Transformation Loss}
Our transformation loss directly optimizes the deviation from estimated and ground truth transformation $R_{pred},t_{pred}$ and $R_{gt},t_{gt}$, that can be calculated as:
\begin{equation}
    \mathcal{L}_{trans}=\varepsilon \Vert R_{gt}^TR_{pred}-I\Vert_1+\Vert t_{gt}-t_{pred}\Vert_2,
\end{equation}
where $\varepsilon$ is a weighting factor.

\subsubsection{Correspondence Distance Loss}
The correspondence distance loss forces the correspondence matrix correctly assign the overlap source points $X'$ to target points $Y'$. Given the ground truth transformation $R_{gt},t_{gt}$, the loss for matrix $C_R$ in rotation branch can be formulated as:
\begin{equation}
    \mathcal{L}_{corr}^R=\frac{1}{|X'|}\Vert R_{gt}X'+t_{gt}-Y'C_R^T\Vert_1.
\end{equation}
The loss in the translation branch is computed in the same way. Total correspondence distance loss is $\mathcal{L}_{corr}=\frac{1}{2}\left(\mathcal{L}_{corr}^R+\mathcal{L}_{corr}^t\right)$.

\subsubsection{Registration Distance Loss}
Our registration distance loss penalizes the $\ell_1$ distance between the source point cloud $X$ transformed using the ground truth $R_{gt}$, $t_{gt}$ and the estimated transformation $R_{pred}$, $t_{pred}$:
\begin{equation}
    \mathcal{L}_{dis}=\frac{1}{|X|}\Vert (R_{gt}X+t_{gt})-(R_{pred}X+t_{pred})\Vert_1.
\end{equation}

The final loss for training the registration module is a weighted sum of three components:
\begin{equation}
    \mathcal{L}_{reg}=\lambda_1\mathcal{L}_{trans}+\lambda_2\mathcal{L}_{corr}+\lambda_3\mathcal{L}_{dis},
\end{equation}
where $\lambda_1$, $\lambda_2$, and $\lambda_3$ are weighting factors.

\section{Experiments}
\subsection{Dataset}
\subsubsection{ModelNet40}
ModelNet40 \cite{wu20153d} is a widely used dataset in the point cloud community. There are 12,311 CAD models from 40 man-made object categories. We uniformly sample 1024 points from each object to construct the source and target point cloud and the two point clouds are sampled separately to fit for purpose. We randomly sample three Euler angle rotations in the range [$-45^{\circ},45^{\circ}$] and translations within [-1,1] on each axis as the rigid transformation. The manner in \cite{yew2020rpm} is applied to generate the partial overlap point cloud.

\subsubsection{3DMatch}
3DMatch \cite{zeng20173dmatch} contains 62 scenes among which 46 are used for training, 8 for validation, and 8 for testing. We use the preprocessed data from Predator that contains voxel-grid downsampled point clouds, the point cloud pairs in the train, valid and test split are 20642, 1331 and 1623 respectively. Since the 3DMatch dataset is converted from real collated RGB-D data so we don’t need to add another operation for the transformation or partial overlap generation.

\subsection{Baseline Methods and Metrics}
We compare our method with traditional methods: ICP \cite{DBLP:journals/pami/BeslM92} and FGR \cite{zhou2016fast} and the learning-based methods: DCP \cite{wang2019deep}, OMNet \cite{xu2021omnet}, RIENet \cite{shen2022reliable}, and FINet \cite{xu2022finet}. Additionally, we compare the effectiveness of our overlap prediction module with MaskNet \cite{sarode2020masknet} and INENet \cite{wu2022inenet}. For ICP and FGR, we use the implementation in Open3D \cite{zhou2018open3d}, and for other algorithms, we use code released by the authors.

Following previous works \cite{wang2019deep,yew2020rpm}, we report root mean squared error (RMSE), mean absolute error (MAE), and isotropic error (Error) for both rotation and translation.

\subsection{Implementation Details}
We conduct experiments with PyTorch \cite{paszke2019pytorch} on an Intel Xeon Glod 6130 CPU and an NVIDIA RTX 2080Ti GPU. The weight factors $\alpha$ and $\gamma$ in overlap loss $\mathcal{L}_o$ are set to 1 and 4 empirically, and the weight factors $\lambda_1, \lambda_2, \lambda_3$ in registration loss $\mathcal{L}_{reg}$ are set to 1, 10, 1 in all experiments. To train the model on ModelNet40, we set weighting factor $\varepsilon $ in $\mathcal{L}_{trans}$ to 1 and overlap threshold to 0.1. To train the model on 3DMatch, we set weighting factor $\varepsilon $ in $\mathcal{L}_{trans}$ to 5 and overlap threshold to 0.0375.

Considering stability our method is trained in two stages. In the first stage, we only train the overlap prediction module with overlap loss $\mathcal{L}_{o}$. We train 300/150 epochs on ModelNet40/3DMatch. In the second stage, we load and froze the weights in the overlap prediction module and optimize the registration part by the aforementioned registration loss $\mathcal{L}_{reg}$. We train 300/50 epochs on ModelNet40/3DMatch. In all experiments, we use the AdamW \cite{DBLP:conf/iclr/LoshchilovH19} optimizer with an initial learning rate of $10^{-4}$ and weight decay of $10^{-4}$ in all experiments and the cosine learning rate schedule with warmup is also applied. For more details please refer to supplementary materials.

\subsection{Evaluation on ModelNet40}
\subsubsection{Unseen Shapes}
In this experiment, we train models on the first 20 categories and evaluate them on the corresponding validation set. Following \cite{yew2020rpm}, we randomly sample a plane and move it to retain 70\% of the point cloud, i.e., 717 out of 1024 points. Table \ref{unseenshapes} shows the results. We can see that our method achieves the best rotation estimation and is on par with FINet in translation estimation. It's worth noting that RIENet fails to obtain reasonable results on our partial overlap setting. For traditional methods, ICP performs poorly because the initial positions of two point clouds vary considerably. FGR outperforms some learning-based methods, however, it uses additional normal data.
\begin{table}[ht]
    \centering
    \caption{Performance on unseen shapes point clouds.}
    \label{unseenshapes}
    \renewcommand\arraystretch{1}
    \resizebox{\columnwidth}{!}{
    % \begin{tabular}{ccccccc}
    %     \hline
    %     Method & RMSE(R) & MAE(R) & RMSE(t) & MAE(t) & Error(R) & Error(t) \\ 
    %     \hline
    %     ICP &  &  &  &  &  &  \\
    %     FGR &  &  &  &  &  &  \\
    %     DCP &  &  &  &  &  &  \\
    %     IDAM &  &  &  &  &  &  \\
    %     OMNet &  &  &  &  &  &  \\
    %     FINet &  &  &  &  &  &  \\ 
    %     Ours &  &  &  &  &  &  \\
    %     \hline
    % \end{tabular}
    
    \begin{tabular}{ccccccc}
        \hline
        \multirow{2}{*}{Method} & \multicolumn{2}{c}{RMSE} & \multicolumn{2}{c}{MAE} & \multicolumn{2}{c}{Error} \\ \cline{2-7} 
         & \multicolumn{1}{c}{R} & \multicolumn{1}{c}{t} & \multicolumn{1}{c}{R} & \multicolumn{1}{c}{t} & \multicolumn{1}{c}{R} & \multicolumn{1}{c}{t} \\ 
        \hline
        ICP & 29.769 & 0.380 & 15.776 & 0.220 & 31.099 & 0.469 \\
        FGR & 21.331 & 0.182 & 4.460 & 0.040 & 7.961 & 0.086 \\
        DCP & 13.650 & 0.218 & 9.693 & 0.157 & 18.895 & 0.321 \\
        OMNet & 8.371 & 0.122 & 3.005 & 0.049 & 6.255 & 0.108 \\
        RIENet & 74.267 & 0.685 & 54.260 & 0.497 & 93.154 & 1.009 \\
        FINet & \underline{6.123} & \textbf{0.078} & \underline{2.329} & \textbf{0.032} & \underline{4.828} & \textbf{0.070} \\ 
        Ours & \textbf{5.381} & \underline{0.081} & \textbf{2.027} & \underline{0.037} & \textbf{4.203} & \underline{0.080} \\
        \hline
    \end{tabular}
    }
    
\end{table}

\subsubsection{Unseen Categories}
To test the generalization capability of our model, we evaluate the performance on unseen categories in this experiment. We use the training set of the first 20 categories to train models and test on the last 20 categories. The results are summarized in Table \ref{unseencategories}. We can find that because of the domain gap between training and testing, all learning-based methods perform worse compared with Table \ref{unseenshapes}. Nevertheless, two traditional algorithms are less affected because they are based on handcrafted optimization objectives, which are insensitive to different input categories. Compared with other methods, our method exhibits excellent robustness and ranks first in all metrics.
\begin{table}[ht]
    \centering
    \caption{Performance on unseen categories point clouds.}
    \label{unseencategories}
    \renewcommand\arraystretch{1}
    \resizebox{\columnwidth}{!}{
    \begin{tabular}{ccccccc}
        \hline
        \multirow{2}{*}{Method} & \multicolumn{2}{c}{RMSE} & \multicolumn{2}{c}{MAE} & \multicolumn{2}{c}{Error} \\ \cline{2-7} 
         & \multicolumn{1}{c}{R} & \multicolumn{1}{c}{t} & \multicolumn{1}{c}{R} & \multicolumn{1}{c}{t} & \multicolumn{1}{c}{R} & \multicolumn{1}{c}{t} \\ 
        \hline
        ICP & 28.464 & 0.375 & 16.333 & 0.222 & 31.543 & 0.466 \\
        FGR & 23.916 & 0.214 & 4.936 & \underline{0.045} & 8.101 & \underline{0.096} \\
        DCP & 16.186 & 0.248 & 11.766 & 0.180 & 22.816 & 0.365 \\
        OMNet & 11.037 & 0.150 & 4.547 & 0.067 & 9.327 & 0.141 \\
        RIENet & 73.727 & 0.682 & 53.772 & 0.495 & 90.670 & 0.998 \\
        FINet & \underline{8.688} & \underline{0.111} & \underline{3.650} & 0.049 & \underline{7.248} & 0.104 \\ 
        Ours & \textbf{6.538} & \textbf{0.090} & \textbf{2.226} & \textbf{0.040} & \textbf{4.541} & \textbf{0.084} \\
        \hline
    \end{tabular}
    }
    
\end{table}

\subsubsection{Gaussian Noise}
\begin{figure}[t]
    \centering
    \includegraphics[width=0.9\columnwidth]{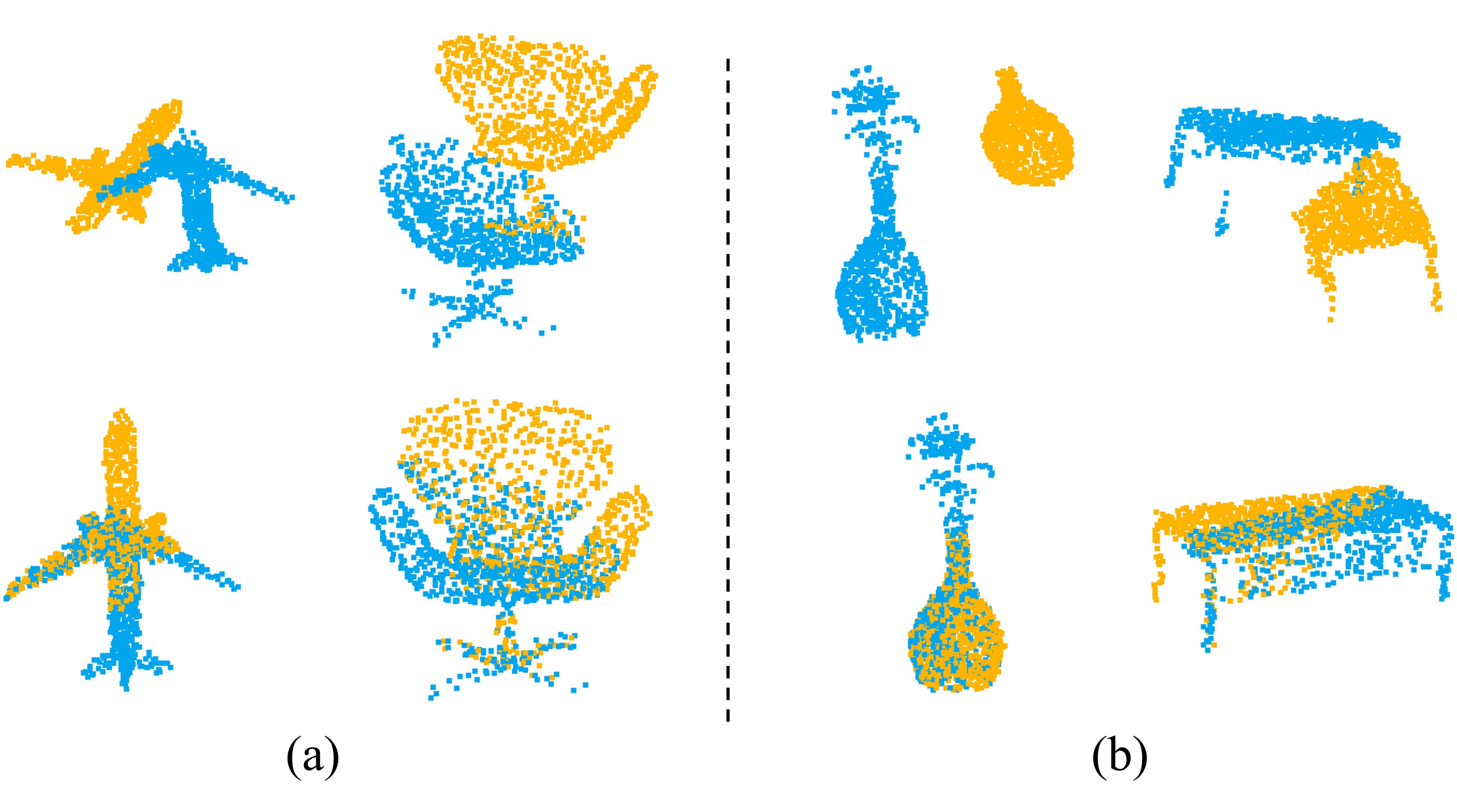} % Reduce the figure size so that it is slightly narrower than the column. Don't use precise values for figure width.This setup will avoid overfull boxes.
    \caption{The qualitative results on Modelnet40 (a) unseen shapes with Gaussian noise, (b) unseen categories with Gaussian noise.}
    \label{modelnet40pic}
\end{figure}

To further test the robustness of our model, we conduct experiments on point clouds contaminated with Gaussian noise. Specifically, we randomly jitter each point in both source and target point clouds by noises sampled from $\mathcal{N}$(0, 0.01) and clipped to [-0.05, 0.05] on each axis. The introduction of noise further reduces the probability that a one-to-one correspondence exists, results for unseen shapes and unseen categories settings are shown in Table \ref{shapesGaussian} and \ref{categoriesGaussian}. FGR achieves decent results on noise-free data, however, it is sensitive to noise, so it performs much worse in this experiment. Compare with Table \ref{unseenshapes} and \ref{unseencategories} we can find that our method is virtually unaffected by noise, which demonstrates the robustness of the proposed DBDNet. In the noise setting, our still exhibits great generalization capability cross domains and achieves the optimum in all measures. Some qualitative results are shown in Fig. \ref{modelnet40pic}.

\begin{table}[ht]
    \centering
    \caption{Performance on unseen shapes with Gaussian noise.}
    \label{shapesGaussian}
    \renewcommand\arraystretch{1}
    \resizebox{\columnwidth}{!}{
    \begin{tabular}{ccccccc}
        \hline
        \multirow{2}{*}{Method} & \multicolumn{2}{c}{RMSE} & \multicolumn{2}{c}{MAE} & \multicolumn{2}{c}{Error} \\ \cline{2-7} 
         & \multicolumn{1}{c}{R} & \multicolumn{1}{c}{t} & \multicolumn{1}{c}{R} & \multicolumn{1}{c}{t} & \multicolumn{1}{c}{R} & \multicolumn{1}{c}{t} \\ 
        \hline
        ICP & 29.950 & 0.391 & 15.881 & 0.229 & 32.100 & 0.482 \\
        FGR & 37.814 & 0.336 & 13.836 & 0.126 & 24.938 & 0.271 \\
        DCP & 14.128 & 0.216 & 10.069 & 0.156 & 19.500 & 0.320 \\
        OMNet & 8.949 & 0.137 & 3.535 & 0.053 & 7.204 & 0.114 \\
        RIENet & 76.470 & 0.716 & 57.029 & 0.528 & 95.947 & 1.068 \\
        FINet & \underline{6.997} & \underline{0.086} & \underline{2.681} & \textbf{0.037} & \underline{5.475} & \textbf{0.078} \\ 
        Ours & \textbf{5.775} & \textbf{0.083} & \textbf{2.231} & \underline{0.040} & \textbf{4.612} & \underline{0.085} \\
        \hline
    \end{tabular}
    }
    
\end{table}
\begin{table}[ht]
    \centering
    \caption{Performance on unseen categories with Gaussian noise.}
    \label{categoriesGaussian}
    \renewcommand\arraystretch{1}
    \resizebox{\columnwidth}{!}{
    \begin{tabular}{ccccccc}
        \hline
        \multirow{2}{*}{Method} & \multicolumn{2}{c}{RMSE} & \multicolumn{2}{c}{MAE} & \multicolumn{2}{c}{Error} \\ \cline{2-7} 
         & \multicolumn{1}{c}{R} & \multicolumn{1}{c}{t} & \multicolumn{1}{c}{R} & \multicolumn{1}{c}{t} & \multicolumn{1}{c}{R} & \multicolumn{1}{c}{t} \\ 
        \hline
        ICP & 27.849 & 0.383 & 16.341 & 0.234 & 32.175 & 0.484 \\
        FGR & 38.140 & 0.302 & 12.739 & 0.102 & 20.569 & 0.218 \\
        DCP & 16.532 & 0.249 & 12.022 & 0.181 & 23.184 & 0.366 \\
        OMNet & 10.500 & 0.163 & 4.661 & 0.078 & 9.472 & 0.162 \\
        RIENet & 71.543 & 0.691 & 52.410 & 0.505 & 89.260 & 1.016 \\
        FINet & \underline{9.001} & \underline{0.120} & \underline{3.827} & \underline{0.055} & \underline{7.748} & \underline{0.117} \\ 
        Ours & \textbf{6.934} & \textbf{0.093} & \textbf{2.368} & \textbf{0.042} & \textbf{4.905} & \textbf{0.088} \\
        \hline
    \end{tabular}
    }
    
\end{table}

\subsection{Evaluation on 3DMatch}
\begin{figure}[t]
    \centering
    \includegraphics[width=0.95\columnwidth]{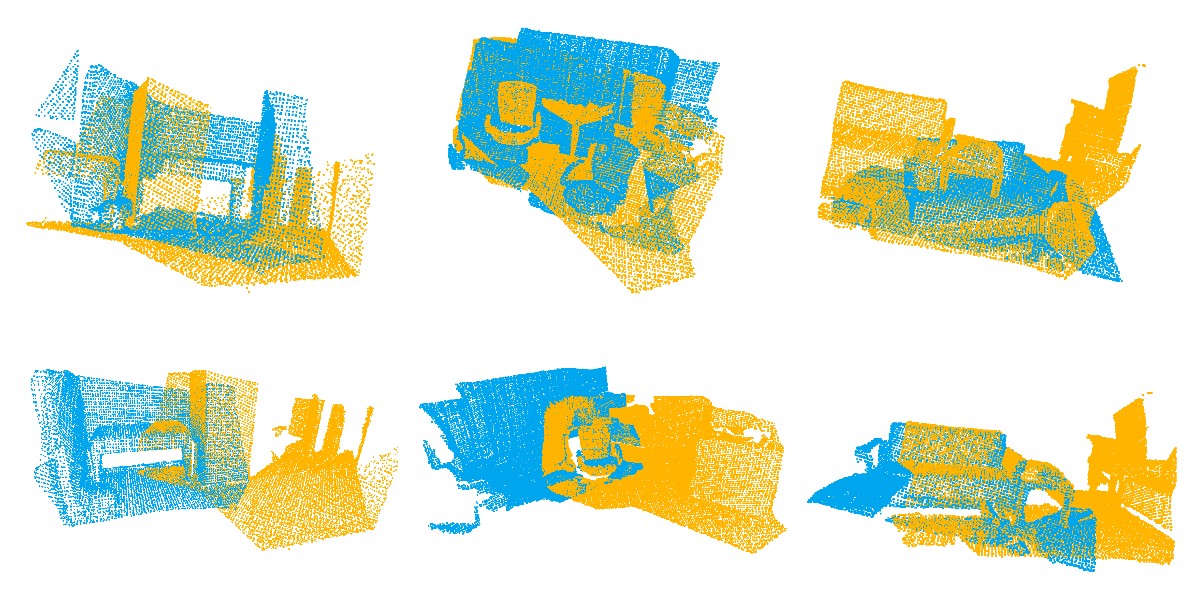} % Reduce the figure size so that it is slightly narrower than the column. Don't use precise values for figure width.This setup will avoid overfull boxes.
    \caption{The qualitative results on 3DMatch.}
    \label{3dmatchpic}
\end{figure}

In this experiment, we evaluate our model using the real indoor dataset 3DMatch. Following \cite{shen2022reliable}, we sample 2048 points to build the source and target point cloud. The results are presented in Table \ref{3dmatch}, demonstrating the significant advantages of our method in terms of both rotation and translation estimation. Some qualitative results are showcased in Fig. \ref{3dmatchpic}.
\begin{table}[ht]
    \centering
    \caption{Performance on 3DMatch.}
    \label{3dmatch}
    \renewcommand\arraystretch{1}
    \resizebox{\columnwidth}{!}{
    \begin{tabular}{ccccccc}
        \hline
        \multirow{2}{*}{Method} & \multicolumn{2}{c}{RMSE} & \multicolumn{2}{c}{MAE} & \multicolumn{2}{c}{Error} \\ \cline{2-7} 
         & \multicolumn{1}{c}{R} & \multicolumn{1}{c}{t} & \multicolumn{1}{c}{R} & \multicolumn{1}{c}{t} & \multicolumn{1}{c}{R} & \multicolumn{1}{c}{t} \\ 
        \hline
        ICP & 33.810 & 0.811 & 21.299 & 0.531 & 37.088 & 1.067 \\
        FGR & 52.333 & 1.072 & 26.907 & 0.568 & 43.005 & 1.141 \\
        DCP & 28.387 & 0.672 & 18.988 & 0.454 & 32.298 & 0.917 \\
        OMNet & \underline{26.255} & \underline{0.611} & \underline{14.219} & \underline{0.371} & \underline{24.007} & \underline{0.752} \\
        RIENet & 53.430 & 1.119 & 34.190 & 0.764 & 59.814 & 1.525 \\
        FINet & 29.515 & 0.736 & 17.929 & 0.471 & 31.360 & 0.946 \\ 
        Ours & \textbf{23.418} & \textbf{0.491} & \textbf{6.701} & \textbf{0.262} & \textbf{8.894} & \textbf{0.570} \\
        \hline
    \end{tabular}
    }
    
\end{table}

\subsection{Evaluation of Overlap Prediction}
In this experiment, we compare our overlap prediction module with two similar works MaskNet \cite{sarode2020masknet} and INENet \cite{wu2022inenet}. We use the ModelNet40 dataset and follow the data configuration from unseen shapes with the Gaussian noise experiment. Since the overlap prediction can be considered as a binary classification task, we measure the standard classification metrics. Table \ref{overlap} shows the results. Benefiting from the explicit information exchange introduced by the powerful attention block, our method achieves superior results in both precision and recall, thus providing tractable point clouds for the registration module.
\begin{table}[ht]
    \centering
    \caption{Performance of overlap prediciton modules.}
    \label{overlap} 
    \renewcommand\arraystretch{1}
    \resizebox{\columnwidth}{!}{
    \begin{tabular}{ccccc}
        \hline
        Method & Accuracy & Precision & Recall & F1-score \\
        \hline
        MaskNet & 0.7894 & 0.8086 & \underline{0.9620} & 0.8786 \\
        INENet & \underline{0.8339} & \underline{0.8530} & 0.9527 & \underline{0.9001} \\
        Ours & \textbf{0.9698} & \textbf{0.9784} & \textbf{0.9833} & \textbf{0.9808} \\
        \hline
    \end{tabular}
    }
    
\end{table}

\subsection{Ablation Studies}
To demonstrate the effectiveness of our components, we use unseen shapes with Gaussian noise setting. Table \ref{ablation} presents a summary of the ablation results. To establish a baseline, we remove all proposed modules and employ an expanded DGCNN \cite{wang2019dynamic}, which is slightly larger than our multi-resolution network, as the feature extractor. Subsequently, we gradually introduce the multi-resolution network (MR), overlap prediction (OP), and dual branches (DB). Additionally, we attempt to jointly predict the overlap mask in the registration module (ID 3 in Table \ref{ablation}). It is evident that all three proposed components yield performance improvements. Specifically, our multi-resolution network outperforms the baseline in rotation estimation while utilizing fewer parameters. Moreover, the inclusion of our overlap prediction module leads to improvements in both rotation and translation performance. The comparison between ID 3 and ID 4 supports our intuition that treating overlap as a preceding task is beneficial. Finally, splitting the rotation and translation calculations into separate branches further enhances the registration results.
\begin{table}[ht]
    \centering
    \caption{Ablation studies. * denotes joint overlap prediction.}
    \label{ablation}
    \renewcommand\arraystretch{1}
    \resizebox{\columnwidth}{!}{
    % \begin{tabular}{c|ccc|cc}
    %     \hline
    %     ID & MR & OP & DB & RMSE(R) & RMSE(t) \\
    %     \hline
    %     1 &  &  &  &  & \\
    %     2 & \checkmark &  &  &  & \\
    %     3 & \checkmark & * &  &  & \\
    %     4 & \checkmark & \checkmark &  &  & \\
    %     5 & \checkmark & \checkmark & \checkmark &  & \\
    %     \hline 
    % \end{tabular}
    \begin{tabular}{c|ccc|cccc}
        \hline
        \multirow{2}{*}{ID} & \multicolumn{3}{c|}{Module} & \multicolumn{2}{c}{RMSE} & \multicolumn{2}{c}{Error} \\ \cline{2-8} 
         & MR & OP & DB & R & t & R & t \\ \hline
        1 &  &  &  & 6.652 & 0.132 & 6.716 & 0.181 \\
        2 & \checkmark &  &  & 6.452 & 0.132 & 6.249 & 0.180 \\
        3 & \checkmark & * &  & 6.603 & 0.107 & 6.350 & 0.134 \\ 
        4 & \checkmark & \checkmark &  & 6.124 & 0.088 & 5.138 & 0.100 \\ 
        5 & \checkmark & \checkmark & \checkmark & \textbf{5.775} & \textbf{0.083} & \textbf{4.612} & \textbf{0.085} \\ 
        \hline
        \end{tabular}
    }
    
\end{table}

\section{Conclusion}
This paper introduces DBDNet, a novel point cloud registration network. Our method utilizes dual branches to separately calculate rotation and translation, effectively eliminating accumulated error. To address partial overlap, we employ an overlap predictor to select overlapping points prior to registration. Additionally, we propose a multi-resolution network for integrating local and global information. Experimental results on scene and object point cloud datasets demonstrate the effectiveness of our proposed method.

\bibliographystyle{IEEEtran}
\bibliography{ref.bib}

\begin{thebibliography}{10}
\providecommand{\url}[1]{#1}
\csname url@samestyle\endcsname
\providecommand{\newblock}{\relax}
\providecommand{\bibinfo}[2]{#2}
\providecommand{\BIBentrySTDinterwordspacing}{\spaceskip=0pt\relax}
\providecommand{\BIBentryALTinterwordstretchfactor}{4}
\providecommand{\BIBentryALTinterwordspacing}{\spaceskip=\fontdimen2\font plus
\BIBentryALTinterwordstretchfactor\fontdimen3\font minus \fontdimen4\font\relax}
\providecommand{\BIBforeignlanguage}[2]{{%
\expandafter\ifx\csname l@#1\endcsname\relax
\typeout{** WARNING: IEEEtran.bst: No hyphenation pattern has been}%
\typeout{** loaded for the language `#1'. Using the pattern for}%
\typeout{** the default language instead.}%
\else
\language=\csname l@#1\endcsname
\fi
#2}}
\providecommand{\BIBdecl}{\relax}
\BIBdecl

\bibitem{arnold2021fast}
E.~Arnold, S.~Mozaffari, and M.~Dianati, ``Fast and robust registration of partially overlapping point clouds,'' \emph{IEEE Robotics and Automation Letters}, vol.~7, no.~2, pp. 1502--1509, 2021.

\bibitem{zhou2021path}
P.~Zhou, R.~Peng, M.~Xu, V.~Wu, and D.~Navarro-Alarcon, ``Path planning with automatic seam extraction over point cloud models for robotic arc welding,'' \emph{IEEE robotics and automation letters}, vol.~6, no.~3, pp. 5002--5009, 2021.

\bibitem{azuma1997survey}
R.~T. Azuma, ``A survey of augmented reality,'' \emph{Presence: teleoperators \& virtual environments}, vol.~6, no.~4, pp. 355--385, 1997.

\bibitem{DBLP:journals/pami/BeslM92}
\BIBentryALTinterwordspacing
P.~J. Besl and N.~D. McKay, ``A method for registration of 3-d shapes,'' \emph{{IEEE} Trans. Pattern Anal. Mach. Intell.}, vol.~14, no.~2, pp. 239--256, 1992. [Online]. Available: \url{https://doi.org/10.1109/34.121791}
\BIBentrySTDinterwordspacing

\bibitem{censi2008icp}
A.~Censi, ``An icp variant using a point-to-line metric,'' in \emph{2008 IEEE International Conference on Robotics and Automation}.\hskip 1em plus 0.5em minus 0.4em\relax Ieee, 2008, pp. 19--25.

\bibitem{yang2015go}
J.~Yang, H.~Li, D.~Campbell, and Y.~Jia, ``Go-icp: A globally optimal solution to 3d icp point-set registration,'' \emph{IEEE transactions on pattern analysis and machine intelligence}, vol.~38, no.~11, pp. 2241--2254, 2015.

\bibitem{wang2019deep}
Y.~Wang and J.~M. Solomon, ``Deep closest point: Learning representations for point cloud registration,'' in \emph{Proceedings of the IEEE/CVF international conference on computer vision}, 2019, pp. 3523--3532.

\bibitem{wang2019prnet}
------, ``Prnet: Self-supervised learning for partial-to-partial registration,'' \emph{Advances in neural information processing systems}, vol.~32, 2019.

\bibitem{yew2020rpm}
Z.~J. Yew and G.~H. Lee, ``Rpm-net: Robust point matching using learned features,'' in \emph{Proceedings of the IEEE/CVF conference on computer vision and pattern recognition}, 2020, pp. 11\,824--11\,833.

\bibitem{qin2022geometric}
Z.~Qin, H.~Yu, C.~Wang, Y.~Guo, Y.~Peng, and K.~Xu, ``Geometric transformer for fast and robust point cloud registration,'' in \emph{Proceedings of the IEEE/CVF conference on computer vision and pattern recognition}, 2022, pp. 11\,143--11\,152.

\bibitem{vaswani2017attention}
A.~Vaswani, N.~Shazeer, N.~Parmar, J.~Uszkoreit, L.~Jones, A.~N. Gomez, {\L}.~Kaiser, and I.~Polosukhin, ``Attention is all you need,'' \emph{Advances in neural information processing systems}, vol.~30, 2017.

\bibitem{huang2022gmf}
X.~Huang, W.~Qu, Y.~Zuo, Y.~Fang, and X.~Zhao, ``Gmf: General multimodal fusion framework for correspondence outlier rejection,'' \emph{IEEE Robotics and Automation Letters}, vol.~7, no.~4, pp. 12\,585--12\,592, 2022.

\bibitem{aoki2019pointnetlk}
Y.~Aoki, H.~Goforth, R.~A. Srivatsan, and S.~Lucey, ``Pointnetlk: Robust \& efficient point cloud registration using pointnet,'' in \emph{Proceedings of the IEEE/CVF conference on computer vision and pattern recognition}, 2019, pp. 7163--7172.

\bibitem{huang2020feature}
X.~Huang, G.~Mei, and J.~Zhang, ``Feature-metric registration: A fast semi-supervised approach for robust point cloud registration without correspondences,'' in \emph{Proceedings of the IEEE/CVF conference on computer vision and pattern recognition}, 2020, pp. 11\,366--11\,374.

\bibitem{wang2022storm}
Y.~Wang, C.~Yan, Y.~Feng, S.~Du, Q.~Dai, and Y.~Gao, ``Storm: Structure-based overlap matching for partial point cloud registration,'' \emph{IEEE Transactions on Pattern Analysis and Machine Intelligence}, vol.~45, no.~1, pp. 1135--1149, 2022.

\bibitem{xu2021omnet}
H.~Xu, S.~Liu, G.~Wang, G.~Liu, and B.~Zeng, ``Omnet: Learning overlapping mask for partial-to-partial point cloud registration,'' in \emph{Proceedings of the IEEE/CVF International Conference on Computer Vision}, 2021, pp. 3132--3141.

\bibitem{huang2021predator}
S.~Huang, Z.~Gojcic, M.~Usvyatsov, A.~Wieser, and K.~Schindler, ``Predator: Registration of 3d point clouds with low overlap,'' in \emph{Proceedings of the IEEE/CVF Conference on computer vision and pattern recognition}, 2021, pp. 4267--4276.

\bibitem{chen2022detarnet}
Z.~Chen, F.~Yang, and W.~Tao, ``Detarnet: Decoupling translation and rotation by siamese network for point cloud registration,'' in \emph{Proceedings of the AAAI Conference on Artificial Intelligence}, vol.~36, no.~1, 2022, pp. 401--409.

\bibitem{schonemann1966generalized}
P.~H. Sch{\"o}nemann, ``A generalized solution of the orthogonal procrustes problem,'' \emph{Psychometrika}, vol.~31, no.~1, pp. 1--10, 1966.

\bibitem{walker1991estimating}
M.~W. Walker, L.~Shao, and R.~A. Volz, ``Estimating 3-d location parameters using dual number quaternions,'' \emph{CVGIP: image understanding}, vol.~54, no.~3, pp. 358--367, 1991.

\bibitem{wang2020deep}
J.~Wang, K.~Sun, T.~Cheng, B.~Jiang, C.~Deng, Y.~Zhao, D.~Liu, Y.~Mu, M.~Tan, X.~Wang \emph{et~al.}, ``Deep high-resolution representation learning for visual recognition,'' \emph{IEEE transactions on pattern analysis and machine intelligence}, vol.~43, no.~10, pp. 3349--3364, 2020.

\bibitem{sarode2020masknet}
V.~Sarode, A.~Dhagat, R.~A. Srivatsan, N.~Zevallos, S.~Lucey, and H.~Choset, ``Masknet: A fully-convolutional network to estimate inlier points,'' in \emph{2020 International Conference on 3D Vision (3DV)}.\hskip 1em plus 0.5em minus 0.4em\relax IEEE, 2020, pp. 1029--1038.

\bibitem{yew2022regtr}
Z.~J. Yew and G.~H. Lee, ``Regtr: End-to-end point cloud correspondences with transformers,'' in \emph{Proceedings of the IEEE/CVF conference on computer vision and pattern recognition}, 2022, pp. 6677--6686.

\bibitem{xu2023point}
J.~Xu, Y.~Zhang, Y.~Zou, and P.~X. Liu, ``Point cloud registration with zero overlap rate and negative overlap rate,'' \emph{IEEE Robotics and Automation Letters}, 2023.

\bibitem{mei2023unsupervised}
G.~Mei, H.~Tang, X.~Huang, W.~Wang, J.~Liu, J.~Zhang, L.~Van~Gool, and Q.~Wu, ``Unsupervised deep probabilistic approach for partial point cloud registration,'' in \emph{Proceedings of the IEEE/CVF Conference on Computer Vision and Pattern Recognition}, 2023, pp. 13\,611--13\,620.

\bibitem{thomas2019delio}
Q.~M. Thomas, O.~Wasenm{\"u}ller, and D.~Stricker, ``Delio: Decoupled lidar odometry,'' in \emph{2019 IEEE Intelligent Vehicles Symposium (IV)}.\hskip 1em plus 0.5em minus 0.4em\relax IEEE, 2019, pp. 1549--1556.

\bibitem{xu2022finet}
H.~Xu, N.~Ye, G.~Liu, B.~Zeng, and S.~Liu, ``Finet: Dual branches feature interaction for partial-to-partial point cloud registration,'' in \emph{Proceedings of the AAAI Conference on Artificial Intelligence}, vol.~36, no.~3, 2022, pp. 2848--2856.

\bibitem{qi2017pointnet}
C.~R. Qi, H.~Su, K.~Mo, and L.~J. Guibas, ``Pointnet: Deep learning on point sets for 3d classification and segmentation,'' in \emph{Proceedings of the IEEE conference on computer vision and pattern recognition}, 2017, pp. 652--660.

\bibitem{qi2017pointnet++}
C.~R. Qi, L.~Yi, H.~Su, and L.~J. Guibas, ``Pointnet++: Deep hierarchical feature learning on point sets in a metric space,'' \emph{Advances in neural information processing systems}, vol.~30, 2017.

\bibitem{wang2019dynamic}
Y.~Wang, Y.~Sun, Z.~Liu, S.~E. Sarma, M.~M. Bronstein, and J.~M. Solomon, ``Dynamic graph cnn for learning on point clouds,'' \emph{ACM Transactions on Graphics (tog)}, vol.~38, no.~5, pp. 1--12, 2019.

\bibitem{wu2019pointconv}
W.~Wu, Z.~Qi, and L.~Fuxin, ``Pointconv: Deep convolutional networks on 3d point clouds,'' in \emph{Proceedings of the IEEE/CVF Conference on computer vision and pattern recognition}, 2019, pp. 9621--9630.

\bibitem{thomas2019kpconv}
H.~Thomas, C.~R. Qi, J.-E. Deschaud, B.~Marcotegui, F.~Goulette, and L.~J. Guibas, ``Kpconv: Flexible and deformable convolution for point clouds,'' in \emph{Proceedings of the IEEE/CVF international conference on computer vision}, 2019, pp. 6411--6420.

\bibitem{zhao2021point}
H.~Zhao, L.~Jiang, J.~Jia, P.~H. Torr, and V.~Koltun, ``Point transformer,'' in \emph{Proceedings of the IEEE/CVF international conference on computer vision}, 2021, pp. 16\,259--16\,268.

\bibitem{DBLP:conf/iclr/MaQYR022}
\BIBentryALTinterwordspacing
X.~Ma, C.~Qin, H.~You, H.~Ran, and Y.~Fu, ``Rethinking network design and local geometry in point cloud: {A} simple residual {MLP} framework,'' in \emph{The Tenth International Conference on Learning Representations, {ICLR} 2022, Virtual Event, April 25-29, 2022}.\hskip 1em plus 0.5em minus 0.4em\relax OpenReview.net, 2022. [Online]. Available: \url{https://openreview.net/forum?id=3Pbra-\_u76D}
\BIBentrySTDinterwordspacing

\bibitem{huang2017densely}
G.~Huang, Z.~Liu, L.~Van Der~Maaten, and K.~Q. Weinberger, ``Densely connected convolutional networks,'' in \emph{Proceedings of the IEEE conference on computer vision and pattern recognition}, 2017, pp. 4700--4708.

\bibitem{he2016deep}
K.~He, X.~Zhang, S.~Ren, and J.~Sun, ``Deep residual learning for image recognition,'' in \emph{Proceedings of the IEEE conference on computer vision and pattern recognition}, 2016, pp. 770--778.

\bibitem{ba2016layer}
J.~L. Ba, J.~R. Kiros, and G.~E. Hinton, ``Layer normalization,'' 2016.

\bibitem{wu2022inenet}
Y.~Wu, Y.~Zhang, X.~Fan, M.~Gong, Q.~Miao, and W.~Ma, ``Inenet: Inliers estimation network with similarity learning for partial overlapping registration,'' \emph{IEEE Transactions on Circuits and Systems for Video Technology}, vol.~33, no.~3, pp. 1413--1426, 2022.

\bibitem{sinkhorn1967concerning}
R.~Sinkhorn and P.~Knopp, ``Concerning nonnegative matrices and doubly stochastic matrices,'' \emph{Pacific Journal of Mathematics}, vol.~21, no.~2, pp. 343--348, 1967.

\bibitem{lin2017focal}
T.-Y. Lin, P.~Goyal, R.~Girshick, K.~He, and P.~Doll{\'a}r, ``Focal loss for dense object detection,'' in \emph{Proceedings of the IEEE international conference on computer vision}, 2017, pp. 2980--2988.

\bibitem{wu20153d}
Z.~Wu, S.~Song, A.~Khosla, F.~Yu, L.~Zhang, X.~Tang, and J.~Xiao, ``3d shapenets: A deep representation for volumetric shapes,'' in \emph{Proceedings of the IEEE conference on computer vision and pattern recognition}, 2015, pp. 1912--1920.

\bibitem{zeng20173dmatch}
A.~Zeng, S.~Song, M.~Nie{\ss}ner, M.~Fisher, J.~Xiao, and T.~Funkhouser, ``3dmatch: Learning local geometric descriptors from rgb-d reconstructions,'' in \emph{Proceedings of the IEEE conference on computer vision and pattern recognition}, 2017, pp. 1802--1811.

\bibitem{zhou2016fast}
Q.-Y. Zhou, J.~Park, and V.~Koltun, ``Fast global registration,'' in \emph{Computer Vision--ECCV 2016: 14th European Conference, Amsterdam, The Netherlands, October 11-14, 2016, Proceedings, Part II 14}.\hskip 1em plus 0.5em minus 0.4em\relax Springer, 2016, pp. 766--782.

\bibitem{shen2022reliable}
Y.~Shen, L.~Hui, H.~Jiang, J.~Xie, and J.~Yang, ``Reliable inlier evaluation for unsupervised point cloud registration,'' in \emph{Proceedings of the AAAI Conference on Artificial Intelligence}, vol.~36, no.~2, 2022, pp. 2198--2206.

\bibitem{zhou2018open3d}
Q.-Y. Zhou, J.~Park, and V.~Koltun, ``Open3d: A modern library for 3d data processing,'' 2018.

\bibitem{paszke2019pytorch}
A.~Paszke, S.~Gross, F.~Massa, A.~Lerer, J.~Bradbury, G.~Chanan, T.~Killeen, Z.~Lin, N.~Gimelshein, L.~Antiga \emph{et~al.}, ``Pytorch: An imperative style, high-performance deep learning library,'' \emph{Advances in neural information processing systems}, vol.~32, 2019.

\bibitem{DBLP:conf/iclr/LoshchilovH19}
\BIBentryALTinterwordspacing
I.~Loshchilov and F.~Hutter, ``Decoupled weight decay regularization,'' in \emph{7th International Conference on Learning Representations, {ICLR} 2019, New Orleans, LA, USA, May 6-9, 2019}.\hskip 1em plus 0.5em minus 0.4em\relax OpenReview.net, 2019. [Online]. Available: \url{https://openreview.net/forum?id=Bkg6RiCqY7}
\BIBentrySTDinterwordspacing

\end{thebibliography}

\end{document}